\documentclass[conference]{IEEEtran}
\usepackage{times}
\usepackage{cite}
\usepackage{graphics}
\usepackage{textcomp}
\usepackage{amsmath,amssymb,amsfonts}
\usepackage{xcolor}
\usepackage{graphicx}
\usepackage{subfigure}
\usepackage{hyperref}
\usepackage{url}
\usepackage[vlined,ruled,linesnumbered]{algorithm2e}
\usepackage{lmodern}
\title{Bandwidth Reduction using Importance Weighted Pruning on Ring AllReduce}

\author{
Zehua Cheng,Zhenghua Xu$^*$\thanks{$^*$Corresponding Author}\\
\textit{State Key Laboratory of Reliability and Intelligence of Electrical Equipment}\\
\textit{Hebei University of Technology}\\
\texttt{limber@snowcloud.ai,zhenghua.xu@hebut.edu.cn}\\
}

\begin{document}
\maketitle

\begin{abstract}
It is inevitable to train large deep learning models on a large-scale cluster equipped with accelerators system. Deep gradient compression would highly increase the bandwidth utilization and speed up the training process but hard to implement on ring structure. In this paper, we find that redundant gradient and gradient staleness has negative effect on training. We have observed that in different epoch and different steps, the neural networks focus on updating different layers and different parameters. In order to save more communication bandwidth and preserve the accuracy on ring structure, which break the restrict as the node increase, we propose a new algorithm to measure the importance of gradients on large-scale cluster implementing ring all-reduce based on the size of the ratio of parameter calculation gradient to parameter value. Our importance weighted pruning approach achieved $64\times$ and $58.8\times$ of gradient compression ratio on AlexNet and ResNet50 on ImageNet. Meanwhile, in order to maintain the sparseness of the gradient propagation, we randomly broadcast the index of important gradients on each node. While the remaining nodes are ready for the index gradient and perform all-reduce update. This would speed up the convergence of the model and preserve the training accuracy.
\end{abstract}

\section{Introduction}
Recently, deep learning have driven tremendous progress in computer vision, natural language processing, and forecasting. In order to have better performance, deeper and bigger models~\cite{szegedy2015going} has trained on larger dataset which large-scale distributed training improves the productivity of training deeper and larger models~\cite{jin2016scale,dean2012large,coates2013deep,chilimbi2014project, shi_modeling_2018}.

However, training on large-scale dataset on distributed system also brought up some split-new problems. The extensive gradients and parameter synchronization extend communication time, which has led to the need for high-end bandwidth solutions for building distributed systems(\cite{li2014scaling,chen2016revisiting,lian2015asynchronous,sra2015adadelay}). However, to improve the training rate, the bandwidth and computing resource utilization are the major challenge for training deep learning model on large-scale distributed system. Synchronous stochastic gradient descent~\cite{das2016distributed} is the most popular approach to scaling ImageNet(\cite{deng2009imagenet}) training. However, to scale synchronous stochastic gradient descent required to increase the batch size requiring faster bandwidth. Many large-scale HPC solutions require Infiniband\cite{pfister2001introduction} to guarantee the communication bandwidth. Deep gradient compression (\cite{lin_deep_2017}) has significantly reduce the communication baxndwidth based on traditional distribution cluster by introducing momentum correction, local gradient clipping, momentum factor masking and warm-up training. However, deep gradient compression adopted on centralized deployment mode which the bottleneck is the high communication cost on the central nodes. \cite{gibiansky2017bringing} has proved that ring all-reduce approach came to have better performance on utilize the bandwidth resources. However, deep gradient compression would restricted by the increasingly dense gradient when the number of training nodes grows on ring structure and sorting according to the size of the gradient results in slower convergence rate. Many methodologies have proved to have weak performance on ring structure which restricted the development of large-scale deep learning training.

In this work, we propose importance weighted pruning on ring all-reduce structure to highly utilize the bandwidth and preserve the training accuracy. We build our large-scale distributed training cluster without using Infiniband and high-end GPUs(we only use NVIDIA GTX 1080ti) which have saved a lot of cost. We have observed that in different epoch and different step, the neural network focuses on updating different layers and different parameters. We further propose layer-wise method to determine the importance of gradient of each layer and send the selected influential gradient to update, leaving gradients accumulation locally to highly utilize the bandwidth. During the training process, as the learning rate continues to decrease, the judgment of the importance of the gradient is getting higher.

\section{Related Work}
Researchers have proposed many approaches to accelerate distributed deep learning training process.  Distributed synchronous Stochastic Gradient Descent (SSGD) is commonly adopted solution for parallelize the tasks across machines. Message exchange algorithm would also speed up training process by make full use of the communication resources. \cite{you2017scaling} utilizing the characteristics of deep learning training process, using different learning rate for different layers based on the norm of the weights($||w||$) and the norm of the gradients($||\nabla w||$) for large batch training. The ratio of weights and gradients $(\frac{||w||}{||\nabla w||})$ varies significantly for different layers. But it is get a equal learning rate on the same layer and it doesn’t distinguish the different learning rate for the same layers because the ratio of weights and gradients ($\frac{||w||}{||\nabla w||}$) varies significantly for same layers.

\textbf{Deep Gradient Compression} Deep gradient compression implementing momentum correction and local gradient pruning on top of gradient sparseness to maintain model performance and save communication bandwidth. However, \cite{lin_deep_2017} copies in computing workers are trained in parallel by applying different subsets of data on centralized parameter server. Centralized distributed cluster restricted by the number of the nodes growth in which case indicated the limit of the training rate. However, implementing deep gradient compression on large-scale ring reduce structure would restricted by the node increases. Because as the number of ring nodes increases, the gradient on each node becomes denser as the ring reduce is performed, so network bandwidth cannot be saved. So if we took the top1\% gradient on each node. As the node passes gradients to another, the worst case is that the top $k$ gradient is 2\%. As the number of nodes increases, the gradient carried by the nodes will continue become denser. Therefore, we believe that the deep gradient compression lost the meaning of spreading the sparse gradient.

\textbf{Gradient Quantization and Sparsification} Quantizing the gradients to low-precision values reduce the communication bandwidth by cutting down the whole model which usually sacrifice a lot of precision ,like \cite{wen2017terngrad}. \cite{seide20141} proposed 1-bit SGD to reduce gradients transfer data size and achieved 10× speedup in traditional speech applications. \cite{strom2015scalable} proposed threshold quantization to only send gradients larger than a predefined constant threshold, but the predefined threshold is hard to determine.

\textbf{Message Exchange Algorithm} Cluster computing resources are restricted by bandwidth and updating strategies. Ring Allreduce(\cite{gibiansky2017bringing}) is an algorithm with constant communication cost. The limit of the system is determined only by the slowest connection between GPUs, which is an ideal solution for large model with large amounts of data exchange. The algorithm proceeds an allgather operation after the scatter-reduce operation. Scatter-reduce operation would exchange the data and every GPU would ends up with a chunk of the final result. Allgather step will exchange those chunks ending up with the complete final result. \cite{zhao2013butterfly} improve the cluster performance with incremental update to solve the communication overhead through the network which has better performance on the traditional machine learning solutions like svm and logistic regression, but models like ResNet\cite{he2016deep} does not have better performance comparing to single ring allreduce approaches.

\section{System Implementation}
\subsection{Ring Structure}
The traditional parameter sever mode has two parts of communication.
The node should first propagate the data to the parameter server to perform operations like sum and average(\cite{li2014scaling,li2014communication}). Then the parameter server broadcasts the processed gradient to the node server. This process limited by the increase of the number of node where both the calculation and communication bandwidth are highly required. The ring structure has less communication operations than the conventional structure.

\begin{figure}[!ht]
\centering
\begin{minipage}[t]{0.48\textwidth}
\centering
\includegraphics[width=6cm]{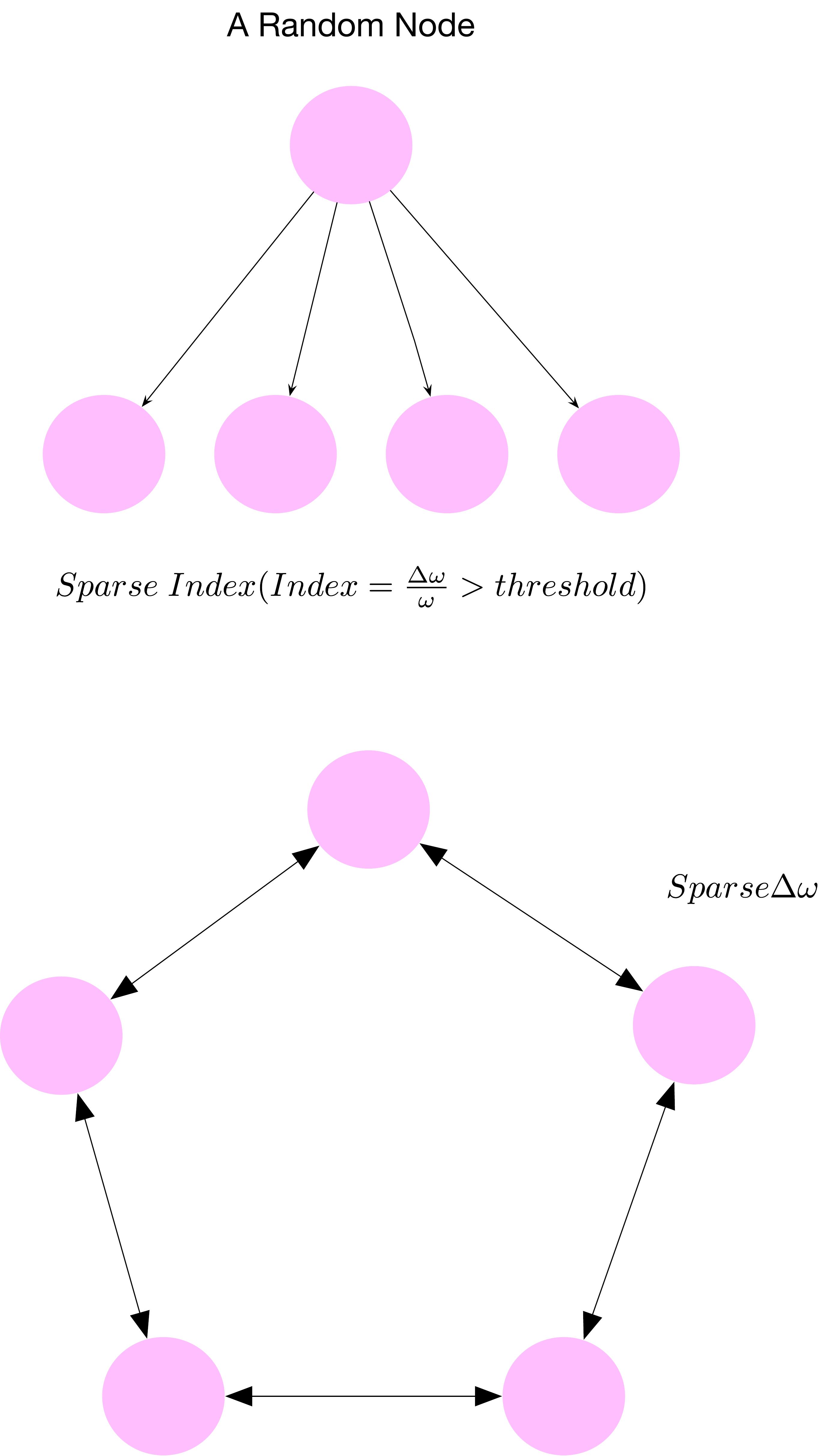}
\caption{Algorithm Topology Structure}\label{figure:algo_topology}
\end{minipage}
\end{figure}

Fig~\ref{figure:algo_topology} has presented the topology of the traditional structure and ring structure. For the ring structure, since there is no central node, we randomly select serveral number of nodes to get the gradient importance to ensure the sparseness of the gradient.

\subsection{Importance Weighted Pruning}
Deep gradient compression(\cite{lin_deep_2017}) has reduce the communication bandwidth by cutting down redundant gradients. \cite{lin_deep_2017} considered the sparse update strategy as important gradients. However, we have observed that the redundant gradients would also make an adverse effort on training, which called the staleness effect. Utilizing the characteristics of deep learning training process seems to have better performance on speed up training rate\cite{you2017scaling}. We solve this problem by sending the important gradients which depending on how much it can change the weight. And we set a threshold to determine which gradients should be transmitted. But a simple important gradient pruning can not adapted to the ring reduce. As the gradient reduce in ring, the sparseness of the gradient will continue to drop. To solve this problem, we randomly choose a node and broadcast its important gradient index, then reduce the important grad on each node.

Distributed training with vanilla momentum SGD on $N$ training nodes follows,
\begin{equation}
    g_{t} = mg_{t-1} + \sum_{k=1}^{N}(\bigtriangledown_{k,t}), \  \   \omega_{t+1} = \omega_{t} -\eta g_{t}
\end{equation}
where $m$ is the momentum, $N$ is the number of training nodes, and $\bigtriangledown_{k,t}\ =\ \frac{1}{NB}\bigtriangledown f(x, \omega_{k,t})$, $B$ is the batchsize, $x$ is the data sampled from $\chi$ at iteration $t$, $\eta$ is the learning rate.

After $T$ iterations, the change for $i$-th layer weight $w^{(i)}$ could be changed as follows,
\begin{equation}
    \omega_{t+T}^{(i)}\ =\ \omega_{t}^{(i)} - \eta (... + (\sum_{\tau =0}^{T-2} m^{\tau})\bigtriangledown_{k, t+1}^{(i)} + (\sum_{\tau =0}^{T-1} m^{\tau})\bigtriangledown_{k, t}^{(i)})
\end{equation}

In our approach, the important gradients will be transmitted and the residual will accumulated local with momentum. So, the momentum SGD with important gradient is changed as follow,
\begin{equation}
    s_{t} = sparse(mg_{t-1} + g_{t}), \ \ \tilde{\omega}_{t+1} = \omega_{t} -\eta s_{t}
\end{equation}

As the update function shows, the parameter becomes differ, but in our algorithm, we choose the gradient which change the parameter  greatly. So we could assume $\omega_{t+1} \approx  \tilde{\omega}_{t+1}$, and this assumption would yield that $s_{t} \approx g_{t}$, follow this assumption, consider the $i$-th layer weight update after T iterations, the change in $\omega^{(i)}$ will as same as in Equation 2.

\begin{algorithm}[!t]
    \begin{small}
    \caption{Importance Weighted Pruning}
    \KwIn{dataset $\chi$\\
    the number of nodes $N$\\
    optimization function SGD\\
    layer parameters $\omega = {\omega [0], \omega [1],...,\omega [M],}$\\
    batch size $B$ per node
    }
    \BlankLine
    \For{$t=1, 2, \dots$}
    {
        $Sample\ data\ x\ from\ \chi$\\
        $\omega^{k} = \omega^{k} + \frac{1}{NB} \bigtriangledown  f(x; \omega _{t})$\\
        \For{$j=0,1,\ddots, M$}
        {$choose\ threshold\ :\ thr$\\
        $choose \ random \ node \ :\ r_{1},\ r_{2},\ddots r_{n}$\\
        $Mask_{r_{i}} \gets \left | \frac{\bigtriangledown \omega_{r}^{k}}{\omega_{r}^{k}} \right | > thr$\\
        $AllGather\ (encode_{uint8}(Mask_{r_{i}}))$\\
        $Mask= \textbf{OR}Mask_{r_i}$\\
        $\tilde{G}_{t}^{k}[j] \gets \bigtriangledown \omega^{k} \odot  Mask$\\
        ${G}_{t}^{k}[j] \gets \bigtriangledown \omega^{k} \odot \lnot Mask $\\
        ${G}_{t}^{k}[j]$ For local accumulation
        }
        $Ring\ all\ reduce$  $\tilde\bigtriangledown \omega^{k} = \frac{\sum \tilde{G}_{t}^{k} }{N}$\\
        ${\omega}_{t+1} \gets SGD( {\omega}_{t}, G_{t})$
        }
    \end{small}
\end{algorithm}

\subsection{Random Gradient Selection}

Gradient staleness can slow down convergence and degrade model performance. In our experiment, most of the parameters are updated between 100-300 steps. The dated gradient will lead to errors in the direction of parameter update. Therefore, the gradient not greater than threshold have a certain probability to update, which can resist the influence of gradient staleness to some extent.

$$P(update)=\frac{gradient\;importance}{threshold}$$

\subsection{Layer-wise Solution}
In large-scale distributed cluster training, we have observed that not every layer contributes the same contribution, which explain the false frozen layer phenomena that after serveral training epochs, the parameters in some particular layer do not update. The mean and variance of the importance of each layer of gradient have certain differences.
Determine the importance of each layer of gradients and send important gradients out for updates. Unimportant gradient local accumulation.

\begin{equation}
threshold=
\left\{
\begin{array}{lr}
    \alpha_{epoch}+\beta_{epoch}\times \frac{var}{mean}(\frac{var}{mean}>C) \\
    \alpha_{epoch}-\beta_{epoch}\times \frac{var}{mean}(\frac{var}{mean}>C)
    \end{array}
    \right.
\end{equation}

\begin{figure}[!ht]
    \centering
    \begin{minipage}[t]{0.48\textwidth}
    \centering
    \includegraphics[width=8cm]{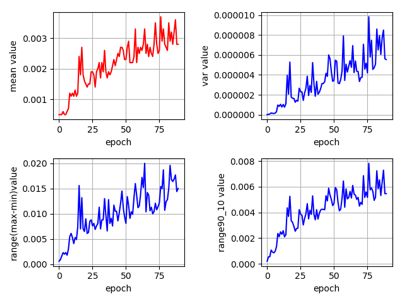}
    \caption{the distribution of gradient importance(convolutional layer)}
    \end{minipage}
    \begin{minipage}[t]{0.48\textwidth}
    \centering
    \includegraphics[width=8cm]{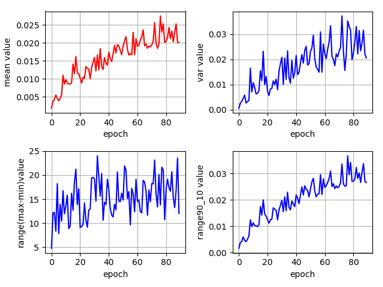}
    \caption{the distribution of gradient importance(batch normalization layer)}
    \end{minipage}
\end{figure}

Choosing $\frac{var}{mean}$ as the factor to hyperparameter to determine the threshold is because we observe that the larger the variance of the gradient importance, the more disordered the gradient and the farther away from the normal distribution, which is not conducive to convergence. so it is necessary to increase the threshold of the gradient importance. When it means very large, the gradient of this layer is very important, and it is beneficial for the update of the parameters, so the threshold needs to be lowered.
At this point, introduce an appropriate threshold can speed up the convergence of the model which determined by mean and variance of the gradient importance.

\begin{figure}[htbp]
\centering
\includegraphics[width=8cm]{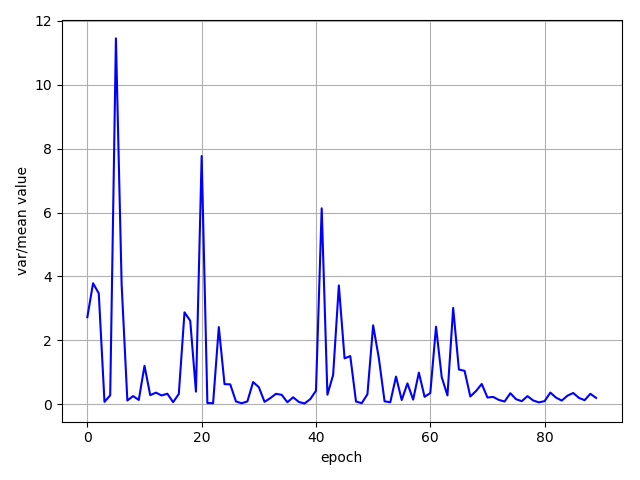}
\caption{The $\frac{var}{mean} of the downsample for the first layer$}
\end{figure}

We can infer that the $\alpha_{epoch}$ hyperparameter is related to epoch at the epoch perspective. $\alpha_{epoch}$ can be set to a constant within a certain epoch interval, for example, $\alpha_{epoch}$ is a constant in epoch 0 to 20 interval. However, epoch 20 variance/mean ratio changes drastically which should slightly adjust the degree of compression. Using $-\beta_{epoch}\times\frac{variance}{mean}(\frac{variance}{mean}<C)$ to cut down the threshold otherwise perform $+\beta_{epoch}\times\frac{variance}{mean}(\frac{variance}{mean}<C)$ to add up the threshold to compress as many gradients as possible.
\section{Experiments}

\subsection{Experiments Settings}
We validate our approach on image classification task. We have tested ResNet50 and AlexNet on ImageNet and ResNet101 on CIFAR10. In this process, we has implemented warm-up training and local gradient clip. In our experiments, we regulated the importance of the threshold in different steps. We have set our threshold to 0.005, 0.01, 0.05, 0.1.
We evaluate the reduction in the network bandwidth by the gradient compression ratio based on the work of \cite{lin_deep_2017}.

$$Gradient Compression Ratio = \frac{size[encode(sparse(G^k))]}{size[G^k]}$$

Because some gradients cannot be updated, in our method, if the importance of the gradient is less than the threshold, we will let the gradient be updated with a certain probability. This probability is positively related to the importance of the gradient.

In our experiments, we have analysis the variance of importance of the hierarchical gradients distribution. In order to prevent the gradient staleness, we determine the probability of updating the gradient

\subsection{Model Result and anlysis}

We have tested resnet50 on ImageNet dataset on our experiment environment which Adopting ring-reduce network topology structure that has 96 nodes. Each node has only one GTX1080ti.

\begin{figure}[htbp]
\centering
\begin{minipage}[t]{0.48\textwidth}
\centering
\includegraphics[width=7cm]{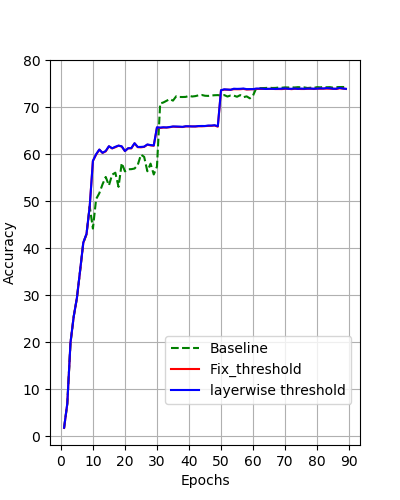}
\caption{Accuracy of ResNet-50 on ImageNet}
\end{minipage}
\begin{minipage}[t]{0.48\textwidth}
\centering
\includegraphics[width=7cm]{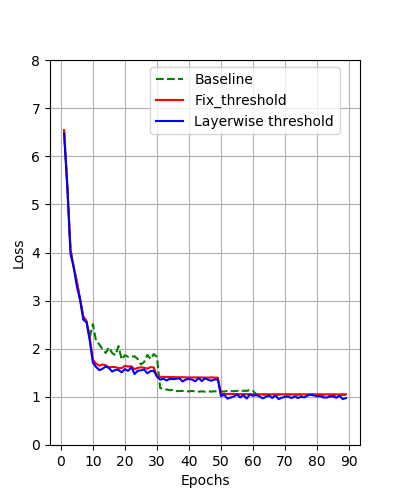}
\caption{Loss of ResNet-50 on ImageNet}
\end{minipage}
\end{figure}

\begin{table}[!ht]
\caption{Comparison of gradient compression ratio on ImageNet Dataset}
\label{sample-table}
\begin{center}
\begin{tabular}{llll}
\multicolumn{1}{c}{Model} &\multicolumn{1}{c}{Training Method}&\multicolumn{1}{c}{Top-1 Accuracy}&\multicolumn{1}{c}{Compress Ratio}
\\ \hline \\
AlexNet          &Baseline &58.17\% &$1\times$ \\
&TernGrad&57.28\%& $8\times$ \\
&Fix Threshold&57.98\%&$\bf{64\times}$\\
&Layerwise Threshold&\textbf{58.19\%}    &$53\times$\\
Resnet50    &Baseline   &76.09\%   &$1\times$\\
            &Fix Threshold  &76.26\%   &$58.8\times$\\
            &\textbf{Layerwise Threshold}&\textbf{76.31\%}    &$\bf{47.6\times}$\\
\end{tabular}
\end{center}
\end{table}

We can observe that our proposed method can guarantee the speed increase under the condition of ensuring a certain precision. In the compression efficiency for the gradient, we can achieve an improvement of about 58 times, and in the case of layering, It can also achieve a 47.6 times improvement.

We also observed that \cite{lin_deep_2017} was performed in our experimental environment, and when performing loop propagation, the original sparse gradient would become denser.

\subsection{Network bandwidth analysis}

\begin{figure*}[!ht]
\begin{center}
\includegraphics[scale=0.4]{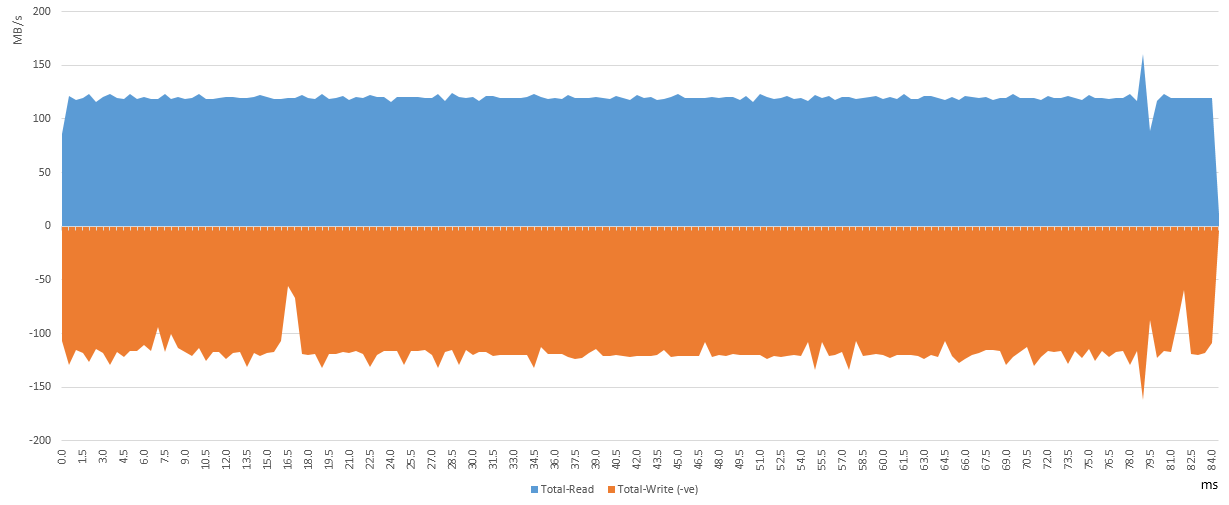}
\end{center}
\caption{Networks I/O for baseline(ResNet50 on ImageNet)(KB/s).}\label{fig:baseline_bandwidth}
\begin{center}
\includegraphics[scale=0.4]{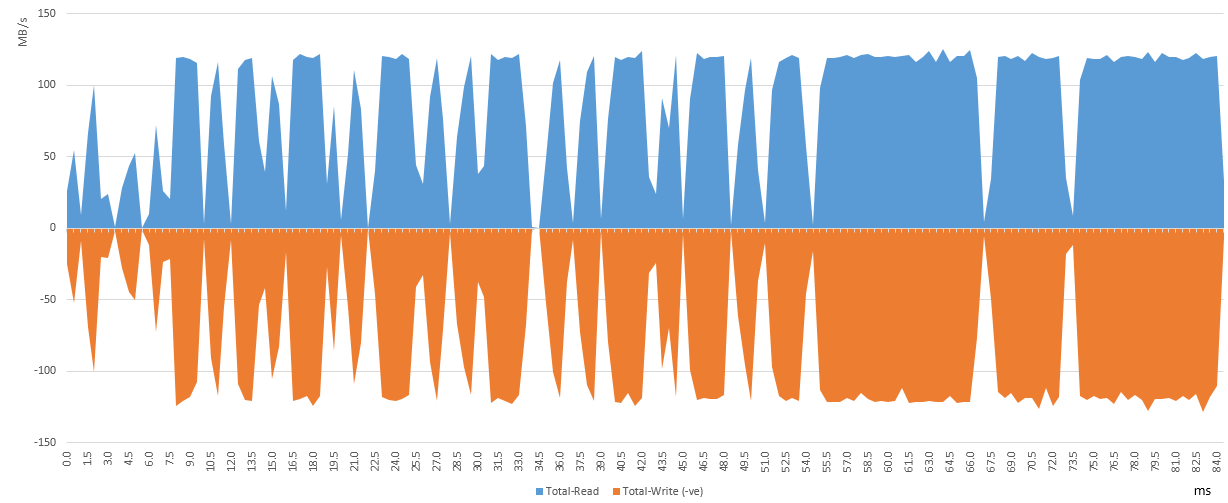}
\end{center}
\caption{Networks I/O after using importance weighted pruning(ResNet50 on ImageNet)(KB/s).}\label{fig:exp_bandwidth}
\end{figure*}

In our approach, we have significantly reduce the bandwidth communication. In Figure~\ref{fig:baseline_bandwidth}, we have observed that in the traditional structure, the Networks I/O bandwidth is close to full load. And easily tell the difference between Figure~\ref{fig:baseline_bandwidth} and Figure~\ref{fig:exp_bandwidth}, we have significantly reduce the Networks I/O.

We can guaranteed the effective communication speed on Gigabit with our importance weighted pruning strategy.

\section{Conclusion}

We propose a new algorithm for measuring the importance of gradients on large-scale cluster implementing ring all-reduce based on the size of the ratio of parameter calculation gradient to parameter value which has speed up the training process and persevering inference accuracy. We made a large-scale deep learning training cluster without using Infiniband. The redundant gradient takes up bandwidth and affect the training process. Our importance weighted pruning on ring all-reduce solution would solve adverse effects of redundant gradient and fully utilize the bandwidth without deficiency when the number of nodes grows.
\bibliographystyle{splncs}
\bibliography{main}

\begin{thebibliography}{10}

\bibitem{szegedy2015going}
Szegedy, C., Liu, W., Jia, Y., Sermanet, P., Reed, S., Anguelov, D., Erhan, D.,
  Vanhoucke, V., Rabinovich, A.:
\newblock Going deeper with convolutions.
\newblock In: Proceedings of the IEEE conference on computer vision and pattern
  recognition. (2015)  1--9

\bibitem{jin2016scale}
Jin, P.H., Yuan, Q., Iandola, F., Keutzer, K.:
\newblock How to scale distributed deep learning?
\newblock arXiv preprint arXiv:1611.04581 (2016)

\bibitem{dean2012large}
Dean, J., Corrado, G., Monga, R., Chen, K., Devin, M., Mao, M., Senior, A.,
  Tucker, P., Yang, K., Le, Q.V.,  et~al.:
\newblock Large scale distributed deep networks.
\newblock In: Advances in neural information processing systems. (2012)
  1223--1231

\bibitem{coates2013deep}
Coates, A., Huval, B., Wang, T., Wu, D., Catanzaro, B., Andrew, N.:
\newblock Deep learning with cots hpc systems.
\newblock In: International Conference on Machine Learning. (2013)  1337--1345

\bibitem{chilimbi2014project}
Chilimbi, T.M., Suzue, Y., Apacible, J., Kalyanaraman, K.:
\newblock Project adam: Building an efficient and scalable deep learning
  training system.
\newblock In: OSDI. Volume~14. (2014)  571--582

\bibitem{shi_modeling_2018}
Shi, S., Wang, Q., Chu, X., Li, B.:
\newblock Modeling and {Evaluation} of {Synchronous} {Stochastic} {Gradient}
  {Descent} in {Distributed} {Deep} {Learning} on {Multiple} {GPUs}.
\newblock arXiv:1805.03812 [cs] (2018) arXiv: 1805.03812.

\bibitem{li2014scaling}
Li, M., Andersen, D.G., Park, J.W., Smola, A.J., Ahmed, A., Josifovski, V.,
  Long, J., Shekita, E.J., Su, B.Y.:
\newblock Scaling distributed machine learning with the parameter server.
\newblock In: OSDI. Volume~14. (2014)  583--598

\bibitem{chen2016revisiting}
Chen, J., Pan, X., Monga, R., Bengio, S., Jozefowicz, R.:
\newblock Revisiting distributed synchronous sgd.
\newblock arXiv preprint arXiv:1604.00981 (2016)

\bibitem{lian2015asynchronous}
Lian, X., Huang, Y., Li, Y., Liu, J.:
\newblock Asynchronous parallel stochastic gradient for nonconvex optimization.
\newblock In: Advances in Neural Information Processing Systems. (2015)
  2737--2745

\bibitem{sra2015adadelay}
Sra, S., Yu, A.W., Li, M., Smola, A.J.:
\newblock Adadelay: Delay adaptive distributed stochastic convex optimization.
\newblock arXiv preprint arXiv:1508.05003 (2015)

\bibitem{das2016distributed}
Das, D., Avancha, S., Mudigere, D., Vaidynathan, K., Sridharan, S., Kalamkar,
  D., Kaul, B., Dubey, P.:
\newblock Distributed deep learning using synchronous stochastic gradient
  descent.
\newblock arXiv preprint arXiv:1602.06709 (2016)

\bibitem{deng2009imagenet}
Deng, J., Dong, W., Socher, R., Li, L.J., Li, K., Fei-Fei, L.:
\newblock Imagenet: A large-scale hierarchical image database.
\newblock In: Computer Vision and Pattern Recognition, 2009. CVPR 2009. IEEE
  Conference on, Ieee (2009)  248--255

\bibitem{pfister2001introduction}
Pfister, G.F.:
\newblock An introduction to the infiniband architecture.
\newblock High Performance Mass Storage and Parallel I/O \textbf{42} (2001)
  617--632

\bibitem{lin_deep_2017}
Lin, Y., Han, S., Mao, H., Wang, Y., Dally, W.J.:
\newblock Deep {Gradient} {Compression}: {Reducing} the {Communication}
  {Bandwidth} for {Distributed} {Training}.
\newblock arXiv:1712.01887 [cs, stat] (December 2017) arXiv: 1712.01887.

\bibitem{gibiansky2017bringing}
Gibiansky, A.:
\newblock Bringing hpc techniques to deep learning.
\newblock Technical report, Baidu Research, Tech. Rep., 2017, http://research.
  baidu. com/bringing-hpc-techniques-deep-learning/. Bingjing Zhang TESTS \&
  CERTIFICATIONS IBM Certified Database Associate-DB2 Universal Database (2017)

\bibitem{you2017scaling}
You, Y., Gitman, I., Ginsburg, B.:
\newblock Scaling sgd batch size to 32k for imagenet training.
\newblock arXiv preprint arXiv:1708.03888 (2017)

\bibitem{wen2017terngrad}
Wen, W., Xu, C., Yan, F., Wu, C., Wang, Y., Chen, Y., Li, H.:
\newblock Terngrad: Ternary gradients to reduce communication in distributed
  deep learning.
\newblock In: Advances in neural information processing systems. (2017)
  1509--1519

\bibitem{seide20141}
Seide, F., Fu, H., Droppo, J., Li, G., Yu, D.:
\newblock 1-bit stochastic gradient descent and its application to
  data-parallel distributed training of speech dnns.
\newblock In: Fifteenth Annual Conference of the International Speech
  Communication Association. (2014)

\bibitem{strom2015scalable}
Strom, N.:
\newblock Scalable distributed dnn training using commodity gpu cloud
  computing.
\newblock In: Sixteenth Annual Conference of the International Speech
  Communication Association. (2015)

\bibitem{zhao2013butterfly}
Zhao, H., Canny, J.:
\newblock Butterfly mixing: Accelerating incremental-update algorithms on
  clusters.
\newblock In: Proceedings of the 2013 SIAM International Conference on Data
  Mining, SIAM (2013)  785--793

\bibitem{he2016deep}
He, K., Zhang, X., Ren, S., Sun, J.:
\newblock Deep residual learning for image recognition.
\newblock In: Proceedings of the IEEE conference on computer vision and pattern
  recognition. (2016)  770--778

\bibitem{li2014communication}
Li, M., Andersen, D.G., Smola, A.J., Yu, K.:
\newblock Communication efficient distributed machine learning with the
  parameter server.
\newblock In: Advances in Neural Information Processing Systems. (2014)  19--27

\end{thebibliography}

\end{document}